\documentclass[11pt,a4paper]{article}
\usepackage[hyperref]{acl2021}
\usepackage{times}
\usepackage{latexsym}
\usepackage{multirow}
\usepackage{amsmath}
\usepackage{amssymb}

\usepackage{color, colortbl}

\definecolor{Gray}{gray}{0.95}
\definecolor{bluegray}{rgb}{0.6784, 0.7529, 0.8078}
\usepackage[first=0,last=9]{lcg}

\usepackage{amssymb}

\usepackage{stfloats}
\usepackage{graphicx}

\usepackage{microtype}

\aclfinalcopy % Uncomment this line for the final submission
\def\aclpaperid{***} %  Enter the acl Paper ID here

\title{On the cross-lingual transferability of multilingual prototypical models across NLU tasks}

\author{Oralie Cattan$^{1,2}$ \And
  Christophe Servan$^{1}$ \And
  Sophie Rosset$^{2}$ \AND \vspace*{-0.5cm}\\ 
  $^{1}$QWANT \\
61 rue de Villiers,\\
92200 Neuilly-sur-Seine, France \\
  \texttt{inital.lastname@qwant.com} \\
  \And \vspace*{-0.5cm}\\
  $^{2}$Université Paris-Saclay, \\CNRS, LISN, \\
  91405, Orsay, France \\
  \texttt{lastname@lisn.fr} \\}

\date{}

\begin{document}
\maketitle
\begin{abstract}
Supervised deep learning-based approaches have been applied to task-oriented dialog and have proven to be effective for limited domain and language applications when a sufficient number of training examples are available. 
In practice, these approaches suffer from the drawbacks of domain-driven design and under-resourced languages.
Domain and language models are supposed to grow and change as the problem space evolves.
On one hand, research on transfer learning has demonstrated the cross-lingual ability of multilingual Transformers-based models to learn semantically rich representations.
On the other, in addition to the above approaches, meta-learning have enabled the development of task and language learning algorithms capable of far generalization. 
Through this context, this article proposes to investigate the cross-lingual transferability of using synergistically few-shot learning with prototypical neural networks and multilingual Transformers-based models.
Experiments in natural language understanding tasks on MultiATIS++ corpus shows that our approach substantially improves the observed transfer learning performances between the low and the high resource languages.
More generally our approach confirms that the meaningful latent space learned in a given language can be can be generalized to unseen and under-resourced ones using meta-learning.

\end{abstract}

\section{Introduction}
%Background
% Task-oriented dialog systems enable humanlike interaction with applications integrated in a broad range of devices, such as smart homes, smartphones or in-car navigation systems.
% By providing a natural language interface for users across various modalities (textual, vocal or visual), they allow a more natural and convenient way to access information and permit to achieve goals over a spectrum of system capabilities. 

% Traditionally, these systems are designed as a pipeline, with a Natural Language Understanding (NLU) as an intermediate module between the user interface and the dialogue management module, which has the role to extract semantic information from a user's query to fill slots in a domain speciﬁc semantic frame.
Traditionally, Natural Language Understanding (NLU) is an intermediate module between the user interface and the dialogue management module in a dialogue system.
It aims to extract semantic information from a user's query or utterance to fill slots in a domain speciﬁc semantic frame.
Domain classification, intent detection and slot filling are three core components belonging to the NLU.
% They are in charge of determining the domain or service of a users query, its underlying goal or intent and associating utterance segments with conceptual labels, called slots, in a manner akin to named entity recognition. 
They are in charge of determining the domain or service of a users query, its underlying goal or intent and associating utterance segments with conceptual labels, called slots, similar to named entity recognition. 

%The Problem: data regime, domain coverage
NLU is usually defined as a supervised learning problem, involving conventional machine learning models on massive amount of annotated training data, which are language dependent.
This prerequisite has prevented its widespread adoption for poorly endowed languages and for small technology companies that do not benefit from millions of users to gather data. 
% Besides the requirement of a large amount of annotated data being available, domains, intents and slots are predefined in taxonomies. %as ontology classes.
Besides the requirement of a large amount of annotated data being available, domains, intents and slots are language dependent. %as ontology classes.
Consequently, in practice, the resulting systems are hardly adaptable to expand to new languages.

As a solution to this problem, cross-lingual transfer approaches were developed to leverage the knowledge from well-resourced languages, with task specific data available to under-resourced languages with little or no data.
%So far, it has been intuitive to apply machine translation to translate the source language into the target language, or vice versa.
%Recently, transfer learning has achieved remarkable progress in various applications.
Recent efforts focused on training Transformer models multilingually such as the multilingual version of BERT \cite{devlin-etal-2019-bert}.
While earlier work demonstrated the effectiveness of multilingual models to learn representations which are transferable across languages, they show limitations when applied to low-resource languages \citep{pires-etal-2019-multilingual, conneau-etal-2020-unsupervised}.
%A survey of this form of learning applied to the field of Natural Language Processing (NLP) can be found in \citet{Ruder2019}.
From another perspective low-shot learning such as few-shot and zero-shot, aims to transfer knowledge learned from one language to another when the training data is limited or is missing some task labels. 

As a core contribution, we explore the potential for cross-lingual transferability of multilingual Transformer-based model \cite{10.5555/3295222.3295349} (mBERT) combined with a few-shot learning algorithm based on prototypical representations. 
We also introduce a zero-shot scenario, where models are trained on multiple languages and evaluated on another.
Our proposed approach relies on appending a mBERT encoder module to the prototypical neural network, which is a proven few-shot model, originally designed for image classification.
Our experimental results show that the generated model trained with a limited number of annotated training examples outperforms the transfer learning based approach on MultiATIS++ dataset \cite{xu-etal-2020-end,upadhyay2018almost} and can be applied to unseen languages directly with decent performance.

\section{Related work}
\label{sec:relwork}

The availability of large datasets has enabled deep learning methods to achieve great success in a variety of fields. 
However, most of these successes are based on supervised learning approaches, which require lots of labeled data to train. 
Most datasets are only available in English. 
Only a few other languages are supported, and most of them are considered as under-resourced languages. 

Recently, meta-learning approaches have enabled the development of task-agnostic learning algorithms capable of far generalizations (cross-domain or cross-lingual) in the context of having a low-data regime. 
Because literature on low-shot learning is vast and diverse, only the most relevant approaches to this work are presented and we refer the reader to \citet{Vanschoren2019} and \citet{10.1145/3293318} for a surveys of earlier work.

\subsection{Low-shot learning}

Humans manifest a capacity of learning new concepts from few stimuli quickly and efficiently by utilizing prior knowledge and experience.
Inspired by this ability, there has been a resurgence of interest in designing specialized models to perform low-shot learning. 
An example of this form of learning is metric-based approaches founded on the simple idea of learning a discriminative metric space in which similar samples are mapped close to each other and dissimilar ones distant.
Siamese \cite{Koch2015SiameseNN}, Matching \cite{10.5555/3157382.3157504} or Prototypical \cite{NIPS2017_6996} networks belong to this category.

\subsubsection{Supervised generalization}

In recent years, several approaches have been introduced and refined to overcome the issue of data-limited regime. %i.e., adapted to the case where a limited amount of labeled data is available.
As an example, the Prototypical Neural Networks (PNNs), developed by \citet{NIPS2017_6996} originally for image classification, were used to extract representative characteristics of the data by mapping data points into an embedding space where each sample will cluster around their respective prototype representation.
\citet{Fort2017} proposed to extend their work by adding a confidence region around prototypes with the help of Gaussian covariance models.%, expressed as a Gaussian covariance matrix.
%and helping to characterize the quality of individual data points.
With the aim of improving the generalization capacity of metric-based methods, \citet{DBLP:journals/corr/abs-1807-02872} proposed to enforce a large margin between the class prototypes by modifying the standard softmax loss function.

\subsubsection{Semi-supervised generalization}

Other approaches, closely related to the aforementioned ones, proposed to take advantage of labeled and unlabeled data. 
Among them, \citet{Boney2017} extended PNNs to address semi-supervised image classification problems. 
They applied a hard clustering to assign the class for the unlabeled examples within the latent space learned by the PNNs.
A close method was developed by \citet{ren2018metalearning} to refine the prototype generation process with clustering.
The authors introduced distractor classes with the aim of handling unlabeled samples not belonging to any of the known classes. 

%Finally, these approaches have mainly been explored in the field of computer vision, as demonstrated in the subsequent sections.
%introduire in NLP
Most of these approaches have mainly been explored in the field of computer vision, and a few of them were applied to NLP fields, such like Natural Language Understanding (NLU).

\subsection{NLU using low-shot learning}

A number of different deep learning approaches have been applied to the problem of language understanding in recent years.
For a thorough overview of deep learning methods in conversational language understanding, we refer the readers to \citet{10.1145/3209978.3210183}.
In the context of relying on limited training resources, few-shot learning has been used for NLU tasks. 
\citet{yazdani-henderson-2015-model} proposes a method to leverage unlabeled data in order to find the separating hyperplanes that divide the utterances with the same label from those with different labels.
\citet{sun-etal-2019-hierarchical} extended PNNs for intent classification using hierarchical attention mechanisms when generating the prototype representations. 
%our approach

Slot filling using few-shot models has also been explored.
\citet{ferreira2015online} presented a zero-shot approach based on a knowledge base and on word representations learned from unlabeled data. 
\citet{10.1145/3297280.3297378} applied PNNs to few-shot named entity recognition by training a separate model for each entity type and \citet{DBLP:journals/corr/abs-1906-08711} proposed a conditional random forest-based approach enhanced with transfer mechanisms that implicitly incorporate label dependencies and similarities.
%our approach
More recently, \citet{dou-etal-2019-investigating}, \citet{bansal-etal-2020-learning} and \citet{bansal-etal-2020-self} applied various meta-learned models to few-shot NLU across domains and tasks.

Finally, besides the approaches of \citet{gu-etal-2018-meta} and \citet{zhang-etal-2020-improving} that focus on handling new and low-resource languages for machine translation, to the best of our knowledge, there are no approaches that combine cross-lingual transfer and meta-learning methods for NLU tasks.

%Focusing on handling new and to adapt to low-resource languages, \citet{gu-etal-2018-meta, zhang-etal-2020-improving} have explored approaches based on multilingual and high-resource language tasks to improve machine translation.

\begin{table*}[h!]
    \centering
    \begin{tabular}{lrrrrrrrrr}
    \hline
\multirow{2}{*}{Language} & \multicolumn{3}{c}{\# utterances} & \multicolumn{3}{c}{\# tokens} & \multirow{2}{*}{\# intents} & \multirow{2}{*}{\# slot types} \\
& \multicolumn{1}{c}{train} & \multicolumn{1}{c}{dev} & \multicolumn{1}{c}{test} & \multicolumn{1}{c}{train} & \multicolumn{1}{c}{dev} & \multicolumn{1}{c}{test} \\
    \hline
English & \multirow{7}{*}{4488} & \multirow{7}{*}{490} & \multirow{7}{*}{893} & 50755 & 5445 & 9164 & \multirow{7}{*}{~~~~~~~~~18} & \multirow{7}{*}{84} \\
Spanish &  &  &  & 55197 & 5927 & 10338 &  &  \\
Portuguese & &  &  & 55052 & 5909 & 10228 &  &  \\
German &  &  &  & 51111 & 5517 & 9383 &  &  \\
French &  &  &  & 55909 & 5769 & 10511 &  &  \\
Chinese &  &  &  & 88194 & 9652 & 16710 &  &  \\
Japanese & \multirow{-7}{*}{\Huge\Bigg\} \normalsize~~~~~~}  &  &  & 133890 & 14416 & 25939 & \multirow{-7}{*}{\Huge\Bigg\} \normalsize~~} &  \\
Hindi & 1440 & 160 & 893 & 16422 & 1753 & 9755 & 17 & 75 \\
Turkish & 578 & 60 & 715 & 6132 & 686 & 7683 & 17 & 71 \\
    \hline
    \end{tabular}
    \caption{Details of the MultiATIS++ corpus.}
    \label{tab:multiatis}
\end{table*}

\section{Approach}

In this section, we present the design of a Prototypical Neural Network and its episodic training procedure before introducing our approach.

\subsection{Prototypical Neural Networks}
\label{sec:proto}

Prototypical Neural Networks \cite{NIPS2017_6996} or PNNs are based on the computation of distance measures between seen-class prototypes to unseen ones. 
More specifically, a $D$-dimensional embedding is generated for each example $x \in $$\mathbb{R}$$^D$ using a neural network based function $f(\cdot)$ parameterized by $\Theta$.
This function enhances the encoding process with better separability properties through a non-linear mapping $f_{\Theta}:  $$\mathbb{R}$$^D \rightarrow $$\mathbb{R}$$^M$. 
The $M$-dimensional prototype of each class is formed as the centroid $c_{i}$ of their embedded support points as seen in Equation \eqref{eq:mean}:

\begin{equation}
\label{eq:mean}
c_{i}=\frac{1}{\left | S_{i} \right |} \sum_{(x_{j}, y_{j}) \in S _{i}}^{ } f_{\Theta}(x_{j}),
\end{equation}

where $S_{i}$ represents the set of examples labeled with class $i$ and $y_{j}$ the corresponding label of $x_{j}$.
%The key assumption is that there exists an embedding in which samples from each class cluster around a single prototype representation of that class.
Equation \eqref{eq:probaquery} shows how, given a query (that is, a new and an unlabeled sample) $q_{i}$, 
the probability distribution over the prototypes is computed from $d(\cdot,\cdot)$, an arbitrary similarity measures function such as the squared euclidean distance or cosine similarity.

\begin{equation}
\label{eq:probaquery}
p _{\Theta}(y_{i}|q_{i})=\frac{exp(- d(f_{\Theta}(q_{i}), c_{i}))}{\sum_{i'}^{} exp(- d(f_{\Theta}(q_{i}), c_{i'}))}
\end{equation}

Finally, the class with the highest probability is chosen by a softmax over the distances and at optimization time, the negative log-probability $J(\Theta)=-\log p _\Theta(y_{i}|q_{i})$ of the true class of each query point is minimized by stochastic gradient descent during an episodic learning process described in the next subsection.

\subsection{Episodic learning}
\label{sec:episodic}

%Few-shot learning is made possible through $N$-way $k$-shot classification tasks using an episodic training procedure \cite{10.5555/3157382.3157504}.
With the aim of generalizing unseen classes from zero to few training examples per class, PNNs is trained from a collection of $N$-way, $k$-shot classification tasks through an episodic training procedure \cite{10.5555/3157382.3157504}.
%It involves a training dataset $D^{train}$ to optimize model parameters and a test dataset $D^{test}$ to evaluate model performance . 
%sampled from a training dataset $D^{train}$ and is evaluated in a similar way on a test dataset $D^{test}$ through an episodic training procedure \cite{10.5555/3157382.3157504}.
Specifically, each episode is one mini-batch consisting of $k$ examples from each of the $N$ classes (both randomly sampled), used to form a labeled (support $S$) and an unlabeled set of examples (query $Q$).
%Each episode is designed to simulate the few-shot task. 
The parameter $k$ often takes a very small value, meaning we have zero-to-$k$ labeled samples.
%The supports are used to construct the class prototypes using Equation \eqref{eq:mean} and the prototypical loss is computed with weight updates based on the query samples according to Equation \eqref{eq:argmax} of Section \ref{sec:proto}
During training, the model is fed with $S$ to construct the class prototypes using Equation \eqref{eq:mean}. 
Its parameters are learned in order to minimize the prototypical loss of its predictions for the examples in the given $Q$ according to Equation \eqref{eq:argmax} of Section \ref{sec:proto}. 
The evaluation is done by averaging the classification performances on query sets of many testing episodes.

\subsection{Transformer-based PNNs}
\label{sec:archi}

%Intent classification tasks are framed as a sequence classification task or sentence-level classification, meaning that the model makes predictions on a single sentence.
%By contrast, slot filling is framed as a sequence labeling task or token-level classification meaning that each word or token (or subword in BERT's case) will be given a label.
%In both cases, we propose an architecture built on top of mBERT model.

Studies have demonstrated that contextualized representations produced by language models such as ELMo \cite{Peters:2018} or BERT \cite{devlin-etal-2019-bert} gave neural networks a better training initializations.
%These models led to a step forward in NLP by deriving more accurate representations.
Rather than training the initialized encoder of PNNs with feature extractors such as convolution or recurrent networks we propose to induce robustness of the pre-trained multilingual BERT (mBERT) to test the distinctiveness of the representation of each class accross languages.
The embedding layer is initialized with the pre-trained mBERT embeddings and fine-tuned together with a dense linear layer that defines the embedding space where the prototype-based classifier operates.
This latent space is used to learn prototypes of each class by estimating their mean and the chosen class is derived from the output layer of the network based on a softmax over distance to the class prototypes.
The motivation behind fine-tuning the encoder with prototypical loss is to induce better generalization properties at test-time to new class labels not seen during training given only a few examples.

\subsection{The cross-lingual way}
\label{sec:cross}

As introduced earlier, even though recent works demonstrate strong cross-lingual transfer capability of multilingual pretrained BERT, they exhibit limitations when applied to low-resource languages \citep{pires-etal-2019-multilingual, conneau-etal-2020-unsupervised}.
To enable cross-lingual transfer according to our few-shot scenario, we construct mutiple episodic batches $E$.
From the available data, we draw the task sets by sampling a subset of labels to form a support set from data in the high-resources languages and a query set from data in the low-resource languages to be evaluated.
NLU data consists of utterances composed of sentence-level intent labels and sequences of slot labels annotated in BIO format \cite{ramshaw-marcus-1995-text} to define the boundary of slots.
The $N$-way $k$-shot NLU task is then defined as follows: 
given an input query utterance in a new language $q_{i}$ and a $k$-shot support set $S$ as references, find the most appropriate intent label or slot label sequence $y$: 

\begin{equation}
\label{eq:argmax}
  \underset{\theta}{argmax_{E}}  \sum_{(q_{i}, y_{i}) \in Q}^{}  \log  p _{\theta}(y_{i}|q_{i}, S).\end{equation}

\section{Experiments}

Our NLU experiments in cross-lingual and few-shot learning for under-resources languages are conducted on MultiATIS++ \cite{xu-etal-2020-end,upadhyay2018almost} corpus, whose description follows.

\subsection{The MultiATIS++ corpus}

MultiATIS++ \cite{upadhyay2018almost, xu-etal-2020-end} is the multilingual extension of the ATIS corpus \cite{hemphill1990atis}, which belongs to the air travel planning domain. 
Originally in English (en), it has been human translated to 8 different other (distant and close) languages i.e., Spanish (es), German (de), French (fr), Portuguese (pt), Hindi (hi), Chinese (zh), Japanese (ja), and Turkish (tr).
It contains 37,084 training examples and 7,859 test examples.
Details of subsets statistics in terms of the number of utterances, intents and slots are shown in Table \ref{tab:multiatis}.
Our main concerns about this corpus are the Hindi and Turkish portions of the data, which are smaller than the other languages, covering only a subset of intents and slots and containing extremely few labeled examples.
%These subsets seem a promising candidate for our few-shot approach, where the scope concerns the language: train on one language and evaluate on another.

%Besides reducing the train set size, we keep the original partitions to compare our results with previous works and take the average scores as final results.

\subsection{Models}

Pour tous les modèles de base construits, nous utilisons les modèles mBERT disponibles au public pré-entraînés sur plus d'une centaine de langages différents \cite{devlin-etal-2019-bert}. 

We use the fine-tuning procedure \cite{devlin-etal-2019-bert} of the original mBERT model as our baseline. 
In sequence-level and token-level classification tasks, it takes the final hidden states (the last layer output of the multi-head Transformer) of the first [CLS] sequence token or each individual token representation as input of the prediction layer to compute classification scores.
Since we plan to use transfer learning in the context of PNNs, we fine-tune the pre-trained mBERT model together with a dense linear layer that defines the embedding space (Section \ref{sec:archi}).

\begin{table*}[h!]
    \centering
    \resizebox{\textwidth}{!}{%
\begin{tabular}{llccccccccc}
\hline
\textbf{config.}                  & \textbf{encoder}       & \textbf{en}                        & \textbf{es}                        & \textbf{de}                        & \textbf{zh}                        & \textbf{ja}                        & \textbf{pt}                        & \textbf{fr}                        & \textbf{hi}                        & \textbf{tr}               \\ \hline
\textbf{target only}              & \textbf{mBERT}         & 98.54                              & \textbf{97.31}                     & 98.43                              & \textbf{97.09}                     & \textbf{97.20}                     & \textbf{97.54}                     & \textbf{98.88}                     & 90.93                              & 83.36            \\
\textbf{}                         & \textbf{mBERT + PNN (5w1s)}  & 97.46                              & 95.14                              & 97.18                              & 96.35                              & 95.53                              & 96.80                               & 97.11                              & 84.95                              & 85.17                     \\
\textbf{}                         & \textbf{mBERT + PNN (5w10s)} & \textbf{98.77}                     & 96.97                              & \textbf{98.54}                     & 97.0                               & 96.64                              & 97.42                              & 97.98                              & \textbf{91.33}                     & \textbf{89.33}            \\
\hline
\textbf{multilingual}             & \textbf{mBERT}         & 98.42                              & 97.98                              & 98.59                              & 97.65                              & 97.45                              & \textbf{98.3}                      & 98.46                              & 95.33                              & \textbf{93.93}            \\
\textbf{}                         & \textbf{mBERT + PNN (5w1s)}  & 95.33                              & 93.71                              & 95.93                              & 95.89                              & 94.42                              & 94.00                               & 94.78                              & 91.4                               & 90.91                     \\
\textbf{}                         & \textbf{mBERT + PNN (5w10s)} & \textbf{99.87}                     & \textbf{98.54}                     & \textbf{98.60}                      & \textbf{98.67}                     & \textbf{98.54}                     & \textbf{98.32}                     & \textbf{98.66}                     & \multicolumn{1}{l}{\textbf{95.49}} & \multicolumn{1}{l}{92.61} \\
\hline
\textbf{multilingual (zero shot)} & \textbf{mBERT}         & \multicolumn{1}{l}{96.42}          & \multicolumn{1}{l}{\textbf{97.98}} & \multicolumn{1}{l}{\textbf{97.54}} & \multicolumn{1}{l}{96.71} & \multicolumn{1}{l}{\textbf{97.45}} & \multicolumn{1}{l}{97.42}          & \multicolumn{1}{l}{\textbf{97.87}} & \multicolumn{1}{l}{\textbf{94.37}} & \multicolumn{1}{l}{\textbf{91.61}} \\
                                  & \textbf{mBERT + PNN (5w1s)}  & \multicolumn{1}{l}{93.73}          & \multicolumn{1}{l}{92.02}          & \multicolumn{1}{l}{93.27}          & \multicolumn{1}{l}{95.62}          & \multicolumn{1}{l}{91.73}          & \multicolumn{1}{l}{93.51}          & \multicolumn{1}{l}{93.28}          & \multicolumn{1}{l}{90.51}          & \multicolumn{1}{l}{89.92} \\
                                  & \textbf{mBERT + PNN (5w10s)} & \multicolumn{1}{l}{\textbf{96.47}} & \multicolumn{1}{l}{97.87}          & \multicolumn{1}{l}{96.86}          & \multicolumn{1}{l}{\textbf{97.65} }         & \multicolumn{1}{l}{96.64}          & \multicolumn{1}{l}{\textbf{98.10}} & \multicolumn{1}{l}{97.45}          & \multicolumn{1}{l}{93.17}          & \multicolumn{1}{l}{90.67} \\ \hline
\end{tabular}
}

    \caption{Averaged intent accuracies obtained with PNNs on 5-way $k$-shot classification $k \in$ [1, 10] (best scores are marked in bold) and baseline results.}
    \label{tab:ic}
\end{table*}

\subsection{Training configurations}

We perform three sets of experiments: \textit{target only}, \textit{multilingual} and \textit{multilingual zero-shot}.

% \begin{itemize}
%     \item \textbf{target only}. This is the configuration consisting of training and testing using only the target language data.
% \end{itemize}

\begin{itemize}
    \item \textbf{target only}: this configuration consists of using only the target language data.
    %, where classifiers are trained from a series of source languages and evaluated on a target language.
\end{itemize}
We also considered two cross-lingual classification tasks with a varying quantity of data between source and target languages to investigate the behaviour of different types of knowledge transfer.
\begin{itemize}
% \begin{itemize}
    \item \textbf{multilingual}: where the training strategy aims to train a network on the concatenation of all of the nine languages and testing the model for each target language.
    \item \textbf{multilingual zero-shot}: where the training relies on the concatenation of all training datasets from all languages except the one we want to test.
\end{itemize}
This works only for the baseline approach (\textit{mBERT}), but with our PNNs approach (\textit{mBERT+PNN}), we performs few-shot learning. This means we use only a few training data in the considered language (\textit{target only} and \textit{multilingual} configurations).

For instance, when we evaluate our approach in the English task, we consider only a fraction of the English training dataset to train our \textit{mBERT+PNN} model in the \textit{target only}. In the \textit{multilingual} configuration, our few-shot approach (\textit{mBERT+PNN}) is trained using only a fraction of all the examples provided for each language.

\subsection{Training details}

For all the baseline models built, we use the publicly available mBERT models pre-trained on over a hundred different languages \cite{devlin-etal-2019-bert}. We trained it using 20 epochs like \citet{xu-etal-2020-end}.

%We use the same encoder for embedding extraction for both support and query points.
%After an experimental phase, 
PNNs training was done using a number of 1000 episodes using Euclidean distance as suggested by the original authors \cite{NIPS2017_6996}.
We consider a configuration parameter and tried a 5-way k-shot intent classification with $k \in$ [1, 10] (5w1s and 5w10s) and 5-way 10-shots slot filling.

For all approaches we use AdamW optimizer \cite{DBLP:journals/corr/abs-1711-05101} using a learning rate of 5e-5 to apply gradients with respect to the loss and weight decay.

All results are reported using the average performances of over 30 runs for intent classification and over 5 runs for slot filling (fewer amount of runs because of higher training time).
%An average pooling operation which has been effective for sentence representation tasks \cite{reimers-gurevych-2019-sentence} was applied to derive sentence embeddings.

\subsection{Results}

Our experimental findings are summarized in Tables \ref{tab:ic} and \ref{tab:sl} for the intent classification and the slot-filling tasks, respectively. 
% Interestingly, we find a great diversity of behaviors across the languages and the approaches.

\subsubsection{Intent classification results}

Using the \textit{target only} configuration, the baseline obtains optimal scores when applied to high resource languages, e.g. \textit{English} (en), \textit{French} (fr) or \textit{German} (de) reaching nearly identical high scores.
%for intention classification and slot filling. 
We obtain the highest baseline scores with an accuracy of 98.8 on the French model, followed by the English model with an accuracy of 98.5.
% and a F1 score of 95.6 on the English model, followed by the French model with an accuracy of 98.8 for intent classification and German for slot filling with a F1 score of 94.88.
Unlike other mainstream languages, the baseline is less accurate on under-resourced languages, with a loss of 7 to 15 points for intent classification on \textit{Hindi} (hi) and \textit{Turkish} (tr) respectively.

In \textit{multilingual} configuration, baseline models perform reasonably well over all the high-resource languages with a significant performance boost due to the availability of additional data. 
The \textit{mBERT + PNN (5w10s)} models outperformed the baseline for all languages, except for the Turkish (tr) language.
% that is most apparent in case of under-resourced languages with a gain of 5 and 10 accuracy points 
% and a gain of 2 and 8 F1 points on \textit{Hindi} (hi) and \textit{Turkish} (tr) languages, respectively.

When transferring from all languages to an unseen one (\textit{multilingual zero-shot} configuration) we observe the best results for the \textit{mBERT} model, except \textit{Portuguese} (pt) and \textit{English} (en) languages, in which the \textit{mBERT + PNN (5w10s)} is 0.5 points better.

Finally, within the framework of the intent classification task, the \textit{mBERT + PNN (5w10s)} model achieves better overall performances in the \textit{multilingual} configuration, especially in the case of under-resourced languages with a gain up to 9 points of accuracy, compared to the \textit{target-only} configuration and an average of one point compared to the best model in the \textit{multilingual zero-shot} configuration.

\begin{table*}[h]
    \centering
    \resizebox{\textwidth}{!}{%
\begin{tabular}{llccccccclc}
\hline
\textbf{config.}                  & \textbf{encoder}             & \textbf{en}                        & \textbf{es}                        & \textbf{de}                        & \textbf{zh}                        & \textbf{ja}                        & \textbf{pt}                        & \textbf{fr}                        & \multicolumn{1}{c}{\textbf{hi}}                  & \textbf{tr}                        \\ \hline
\textbf{target only}              & \textbf{mBERT}               & { 95.64}       & { 85.52}       & { 94.88}       & { 92.93}       & { 93.13}       & { 91.71}       & { 92.78}       & \multicolumn{1}{c}{{ 85.12}} & { 78.22}       \\
\textbf{}                         & \textbf{mBERT + PNN (5w10s)} & \textbf{95.76}                     & \textbf{87.40}                      & \textbf{95.63}                             & \textbf{93.45}                     & \textbf{93.93}                     & \textbf{92.22}                     & \textbf{93.13}                     & \multicolumn{1}{c}{\textbf{85.70}}               & \textbf{82.67}                     \\
\hline
\textbf{multilingual}             & \textbf{mBERT}               & 96.02                              & 88.03                              & 95.03                              & 93.63                              & 93.01                              & 92.31                              & 91.18                              & 87.39                                   & \multicolumn{1}{l}{86.83}          \\
\textbf{}                         & \textbf{mBERT + PNN (5w10s)} & \textbf{98.40}                      & \textbf{92.09}                     & \textbf{97.12}                     & \textbf{95.50}                     & \textbf{97.24}                     & \textbf{95.81}                     & \multicolumn{1}{l}{\textbf{96.80}} & \textbf{89.59}                                           & \textbf{88.39}                     \\
\hline
\textbf{multilingual (zero shot)} & \textbf{mBERT}               & \multicolumn{1}{l}{\textbf{94.10}}           & \multicolumn{1}{l}{\textbf{87.14}}          & \multicolumn{1}{l}{\textbf{94.23}} & \multicolumn{1}{l}{\textbf{92.17}} & \multicolumn{1}{l}{\textbf{92.61}} & \multicolumn{1}{l}{\textbf{91.59}} & \multicolumn{1}{l}{\textbf{90.79}} & 86.14                                   & \multicolumn{1}{l}{85.86}          \\
                                  & \textbf{mBERT + PNN (5w10s)} & \multicolumn{1}{l}{93.25} & \multicolumn{1}{l}{86.99} & \multicolumn{1}{l}{93.57}          & \multicolumn{1}{l}{91.82}          & \multicolumn{1}{l}{92.38}          & \multicolumn{1}{l}{91.19}          & \multicolumn{1}{l}{90.39}          & \textbf{87.49}                                            & \multicolumn{1}{l}{\textbf{86.83}} \\ \hline
\end{tabular}

}

    \caption{Averaged slot F1s obtained with PNNs on 5-way 10-shot and baseline results (highest scores are marked in bold).}
    \label{tab:sl}
\end{table*}

\subsubsection{Slot-filling results}

Slot-filling result trends in the \textit{target only} configuration are about one point better of F1 score for the \textit{mBERT + PNN (5w10s)} model compared to the baseline model (\textit{mBERT}). 
The \textit{mBERT + PNN (5w10s)} model even outperformed the baseline by more than 4 points of F1 in the Turkish task (tr).

We can observe the same trend in the  \textit{multilingual} configuration: our approach outperformed the baseline in all languages.

On the contrary, the \textit{mBERT + PNN (5w10s)} fails in most of language tasks in the \textit{multilingual zero-shot} configuration, except for the \textit{Hindi} (hi) and the \textit{Turkish} (tr) languages.

Finally, like the intent classification task, the \textit{mBERT + PNN (5w10s)} model achieves better overall performance in the \textit{multilingual} configuration for all languages.
% , especially in the case of under-resourced languages with a gain of 5 and 10 accuracy points, compared to the \textit{target-only} configuration and an average of one point compared to the best model in the \textit{multilingual zero-shot} configuration.

\subsection{Result analysis}

% We organize the discussion of our results around questions related to the cross-lingual transferability of mBERT-based model between low- and high-resource languages.

First, our baseline results are on par with those obtained by \citet{qin-etal-2019-stack} and  \citet{xu-etal-2020-end} when they trained BERT-based models using only English training data (en) with intent accuracy scores of 97.5\% and 96.08\% while we obtain 98.5\%. This is the same in our slot-filling experiment in which they report 94.7 F1 points while we obtain 95.6. This difference comes from our results averaging between 30 and 5 runs for intent classification and slot filling, while previous works only performed 5 runs.
We also observe that, just like \citet{xu-etal-2020-end}, slot filling on \textit{Spanish} (es) leads to lower results, similar to those obtained in our few-shot setting.

When transferring from all languages to an unseen one (\textit{multilingual zero-shot configuration} in both tables \ref{tab:ic} and \ref{tab:sl}) we obtained lower scores than the \textit{multilingual} configurations. 
This means the multilingual representation captured in mBERT is efficient enough when data is available in several languages and none are available in the target considered language.
%The performances are still appropriate if we the lack of training data in the language considered for the evaluation.
But, in both cases, the combination of mBERT$+$PNN performs better when fewer data is available using the few-shot approach (the \textit{multilingual} configuration). 
This means that our approach quickly adapts to the considered target language with only a few examples available and enhances the mBERT multilingual transfer learning capabilities. 
This is especially true in the case of slot filling with gains in terms of F1-scores ranging from 2 to 5 points.
% This combination 
% This also confirms the need of monolingual and additional data from other languages to improve system performance. 
% In both intention classification and slot-filling tasks, our few-shot approach based on PNNs outperformed the baseline approach in the \textit{multilingual} configuration. 
% This means that our approach quickly adapts to the considered target language with only a few examples and enhances the mBERT transfer learning capabilities. 
% This is especially true in the case of slot filling with gains in terms of F1-scores ranging from 2 to 5 points.

Finally, using the mBERT baseline model, transfer learning to \textit{French} or \textit{German} has performance scores similar to \textit{English} while using the \textit{Turkish} (tr) or \textit{Hindi} (hi) yielded significant loss.
This leads us to the same conclusion as \citet{xu-etal-2020-end}: exploiting language interrelationships learnt with transfer learning improve the model performances.
This may come from the fact that French, English and German are similar and share some vocabulary while Turkish or Hindi are dissimilar to European languages \cite{hock2019language}.
% This illustrates the performance gains that can be achieved by exploiting language interrelationships learnt with transfer learning.
% , a conclusion further emphasised by the fact that multilingual results outperformed the other configuration models (target only and specifically multilingual zero-shot) regardless of the approach.

% and completely fail, losing up to 17 points for slot filling on \textit{Hindi} (hi). 
% We can notice that just like \citet{xu-etal-2020-end} slot filling on Spanish (es) lead to low results, similar to those obtained in low-resource settings with a slot F1 at 85.52.

% As the results show, PNNs models outperform mBERT-based transfer learning approach.
% In one-shot intent classification experiments (5w1s), the most difficult of cases due to only having one training example per class, PNNs lead to lower results than training using 10 examples (5w10s).
% Finally, it is remarkable that our few-shot approach outperforms with a large margin of the results obtained from the baseline, especially for the slot filling task with gains ranging from 1 to 5 points.

A detailed inspection of the PNNs results shows that in the \textit{target only} and in the \textit{multilingual} configurations, there is an overall and important reduction in recall values, which is balanced by an improvement of the precision values.
If we analyze deeper the mislabeled examples we can observe that applying PNNs help to prevent overlapping and annotation mismatch cases that occur in the data.

We observed that MultiAtis++ corpus seems to be a highly unbalanced labeled dataset with the number of training examples per class varying from 1 to 3300. 
This impacts the model performance, and it could explain why we observe a lower recall and an  improvement in precision using our approach, since it is based on the reduction of the amount of data. 
%In this sense, our approach seems to rebalance the dataset which is promising.
%Considering this, since 
% since our approach is based on the reduction of the amount of data, it seems to have an impact on the representativity allows to increase the balance between classes.
% In this sense, our results seem to be efficient and very promising.

\section{Conclusions}

In this paper, we demonstrate the opportunities in leveraging mBERT-based modeling using few-shot learning for both intent classification and slot filling tasks on under-resource languages. 
% We found that our approach models is a highly effective technique for training low-resource models for closely related languages. 
We found that our approach model is a highly effective technique for training models for low-resource languages. 
% Using the mBERT baseline model, transfer learning to French or German has performance scores similar to English while using the Turkish or Hindi yielded significant loss.
% Since French and German are similar to English while  Turkish or Hindi are distant.
This illustrates the performance gains that can be achieved by exploiting language interrelationships learnt with transfer learning, a conclusion further emphasised by the fact that multilingual results outperformed other configuration models (target only and specifically multilingual zero-shot) regardless of the approach.
Overall, PNNs models outperform mBERT-based transfer learning approach, enabling us to train competitive NLU systems for under-resources languages with only a fraction of training examples.

From this work a new challenge naturally comes up and a possible direction is to adapt a few-shot setting to a joint approach of intent detection and slot filling, like in \citet{10.5555/3060832.3061040}, \citet{liu2016attention-based} and \citet{zhang-etal-2019-joint}, which  demonstrates that performing these two tasks jointly improves the performance of both of them.

\bibliographystyle{acl_natbib}
\bibliography{anthology,acl2021}

\begin{thebibliography}{35}
\expandafter\ifx\csname natexlab\endcsname\relax\def\natexlab#1{#1}\fi

\bibitem[{Bansal et~al.(2020{\natexlab{a}})Bansal, Jha, and
  McCallum}]{bansal-etal-2020-learning}
Trapit Bansal, Rishikesh Jha, and Andrew McCallum. 2020{\natexlab{a}}.
\newblock \href {https://www.aclweb.org/anthology/2020.coling-main.448}
  {Learning to few-shot learn across diverse natural language classification
  tasks}.
\newblock In \emph{Proceedings of the 28th International Conference on
  Computational Linguistics}, pages 5108--5123, Barcelona, Spain (Online).
  International Committee on Computational Linguistics.

\bibitem[{Bansal et~al.(2020{\natexlab{b}})Bansal, Jha, Munkhdalai, and
  McCallum}]{bansal-etal-2020-self}
Trapit Bansal, Rishikesh Jha, Tsendsuren Munkhdalai, and Andrew McCallum.
  2020{\natexlab{b}}.
\newblock \href {https://doi.org/10.18653/v1/2020.emnlp-main.38}
  {Self-supervised meta-learning for few-shot natural language classification
  tasks}.
\newblock In \emph{Proceedings of the 2020 Conference on Empirical Methods in
  Natural Language Processing (EMNLP)}, pages 522--534, Online. Association for
  Computational Linguistics.

\bibitem[{Boney and Ilin(2017)}]{Boney2017}
Rinu Boney and Alexander Ilin. 2017.
\newblock Semi-supervised few-shot learning with prototypical networks.
\newblock In \emph{Workshop on Meta-Learning 2017 (NIPS 2017)}.

\bibitem[{Conneau et~al.(2020)Conneau, Khandelwal, Goyal, Chaudhary, Wenzek,
  Guzm{\'a}n, Grave, Ott, Zettlemoyer, and
  Stoyanov}]{conneau-etal-2020-unsupervised}
Alexis Conneau, Kartikay Khandelwal, Naman Goyal, Vishrav Chaudhary, Guillaume
  Wenzek, Francisco Guzm{\'a}n, Edouard Grave, Myle Ott, Luke Zettlemoyer, and
  Veselin Stoyanov. 2020.
\newblock \href {https://doi.org/10.18653/v1/2020.acl-main.747} {Unsupervised
  cross-lingual representation learning at scale}.
\newblock In \emph{Proceedings of the 58th Annual Meeting of the Association
  for Computational Linguistics}, pages 8440--8451, Online. Association for
  Computational Linguistics.

\bibitem[{Devlin et~al.(2019)Devlin, Chang, Lee, and
  Toutanova}]{devlin-etal-2019-bert}
Jacob Devlin, Ming-Wei Chang, Kenton Lee, and Kristina Toutanova. 2019.
\newblock \href {https://doi.org/10.18653/v1/N19-1423} {{BERT}: Pre-training of
  deep bidirectional transformers for language understanding}.
\newblock In \emph{Proceedings of the 2019 Conference of the North {A}merican
  Chapter of the Association for Computational Linguistics: Human Language
  Technologies, Volume 1 (Long and Short Papers)}, pages 4171--4186,
  Minneapolis, Minnesota. Association for Computational Linguistics.

\bibitem[{Dou et~al.(2019)Dou, Yu, and
  Anastasopoulos}]{dou-etal-2019-investigating}
Zi-Yi Dou, Keyi Yu, and Antonios Anastasopoulos. 2019.
\newblock \href {https://doi.org/10.18653/v1/D19-1112} {Investigating
  meta-learning algorithms for low-resource natural language understanding
  tasks}.
\newblock In \emph{Proceedings of the 2019 Conference on Empirical Methods in
  Natural Language Processing and the 9th International Joint Conference on
  Natural Language Processing (EMNLP-IJCNLP)}, pages 1192--1197, Hong Kong,
  China. Association for Computational Linguistics.

\bibitem[{Ferreira et~al.(2015)Ferreira, Jabaian, and
  Lefèvre}]{ferreira2015online}
Emmanuel Ferreira, Bassam Jabaian, and Fabrice Lefèvre. 2015.
\newblock Online adaptative zero-shot learning spoken language understanding
  using word-embedding.
\newblock \emph{IEEE International Conference on Acoustics, Speech and SP}.

\bibitem[{Fort(2017)}]{Fort2017}
Stanislav Fort. 2017.
\newblock Gaussian prototypical networks for few-shot learning on omniglot.
\newblock In \emph{Workshop on Bayesian Deep Learning (NIPS 2017)}.

\bibitem[{Fritzler et~al.(2019)Fritzler, Logacheva, and
  Kretov}]{10.1145/3297280.3297378}
Alexander Fritzler, Varvara Logacheva, and Maksim Kretov. 2019.
\newblock \href {https://doi.org/10.1145/3297280.3297378} {Few-shot
  classification in named entity recognition task}.
\newblock In \emph{Proceedings of the 34th ACM/SIGAPP Symposium on Applied
  Computing}, SAC ’19, page 993–1000, New York, NY, USA. Association for
  Computing Machinery.

\bibitem[{Gao et~al.(2018)Gao, Galley, and Li}]{10.1145/3209978.3210183}
Jianfeng Gao, Michel Galley, and Lihong Li. 2018.
\newblock \href {https://doi.org/10.1145/3209978.3210183} {Neural approaches to
  conversational ai}.
\newblock In \emph{The 41st International ACM SIGIR Conference on Research and
  Development in Information Retrieval}, SIGIR ’18, page 1371–1374, New
  York, NY, USA. Association for Computing Machinery.

\bibitem[{Gu et~al.(2018)Gu, Wang, Chen, Li, and Cho}]{gu-etal-2018-meta}
Jiatao Gu, Yong Wang, Yun Chen, Victor O.~K. Li, and Kyunghyun Cho. 2018.
\newblock \href {https://doi.org/10.18653/v1/D18-1398} {Meta-learning for
  low-resource neural machine translation}.
\newblock In \emph{Proceedings of the 2018 Conference on Empirical Methods in
  Natural Language Processing}, pages 3622--3631, Brussels, Belgium.
  Association for Computational Linguistics.

\bibitem[{Hemphill et~al.(1990)Hemphill, Godfrey, and
  Doddington}]{hemphill1990atis}
Charles~T Hemphill, John~J Godfrey, and George~R Doddington. 1990.
\newblock The atis spoken language systems pilot corpus.
\newblock In \emph{Speech and Natural Language: Proceedings of a Workshop Held
  at Hidden Valley, Pennsylvania, June 24-27, 1990}.

\bibitem[{Hock and Joseph(2019)}]{hock2019language}
Hans~Henrich Hock and Brian~D Joseph. 2019.
\newblock \emph{Language history, language change, and language relationship:
  An introduction to historical and comparative linguistics}.
\newblock Walter de Gruyter GmbH \& Co KG.

\bibitem[{Hou et~al.(2019)Hou, Zhou, Liu, Wang, Che, Liu, and
  Liu}]{DBLP:journals/corr/abs-1906-08711}
Yutai Hou, Zhihan Zhou, Yijia Liu, Ning Wang, Wanxiang Che, Han Liu, and Ting
  Liu. 2019.
\newblock \href {http://arxiv.org/abs/1906.08711} {Few-shot sequence labeling
  with label dependency transfer}.
\newblock \emph{CoRR}, abs/1906.08711.

\bibitem[{Koch(2015)}]{Koch2015SiameseNN}
Gregory~R. Koch. 2015.
\newblock Siamese neural networks for one-shot image recognition.
\newblock In \emph{ICML Deep Learning Workshop}.

\bibitem[{Liu and Lane(2016)}]{liu2016attention-based}
Bing Liu and Ian Lane. 2016.
\newblock Attention-based recurrent neural network models for joint intent
  detection and slot filling.
\newblock \emph{INTERSPEECH}, pages 685--689.

\bibitem[{Loshchilov and Hutter(2017)}]{DBLP:journals/corr/abs-1711-05101}
Ilya Loshchilov and Frank Hutter. 2017.
\newblock \href {http://arxiv.org/abs/1711.05101} {Fixing weight decay
  regularization in adam}.
\newblock \emph{CoRR}, abs/1711.05101.

\bibitem[{Peters et~al.(2018)Peters, Neumann, Iyyer, Gardner, Clark, Lee, and
  Zettlemoyer}]{Peters:2018}
Matthew~E. Peters, Mark Neumann, Mohit Iyyer, Matt Gardner, Christopher Clark,
  Kenton Lee, and Luke Zettlemoyer. 2018.
\newblock Deep contextualized word representations.
\newblock In \emph{Proc. of NAACL}.

\bibitem[{Pires et~al.(2019)Pires, Schlinger, and
  Garrette}]{pires-etal-2019-multilingual}
Telmo Pires, Eva Schlinger, and Dan Garrette. 2019.
\newblock \href {https://doi.org/10.18653/v1/P19-1493} {How multilingual is
  multilingual {BERT}?}
\newblock In \emph{Proceedings of the 57th Annual Meeting of the Association
  for Computational Linguistics}, pages 4996--5001, Florence, Italy.
  Association for Computational Linguistics.

\bibitem[{Qin et~al.(2019)Qin, Che, Li, Wen, and Liu}]{qin-etal-2019-stack}
Libo Qin, Wanxiang Che, Yangming Li, Haoyang Wen, and Ting Liu. 2019.
\newblock \href {https://doi.org/10.18653/v1/D19-1214} {A stack-propagation
  framework with token-level intent detection for spoken language
  understanding}.
\newblock In \emph{Proceedings of the 2019 Conference on Empirical Methods in
  Natural Language Processing and the 9th International Joint Conference on
  Natural Language Processing (EMNLP-IJCNLP)}, pages 2078--2087, Hong Kong,
  China. Association for Computational Linguistics.

\bibitem[{Ramshaw and Marcus(1995)}]{ramshaw-marcus-1995-text}
Lance Ramshaw and Mitch Marcus. 1995.
\newblock \href {https://www.aclweb.org/anthology/W95-0107} {Text chunking
  using transformation-based learning}.
\newblock In \emph{Third Workshop on Very Large Corpora}.

\bibitem[{Ren et~al.(2018)Ren, Ravi, Triantafillou, Snell, Swersky, Tenenbaum,
  Larochelle, and Zemel}]{ren2018metalearning}
Mengye Ren, Sachin Ravi, Eleni Triantafillou, Jake Snell, Kevin Swersky,
  Josh~B. Tenenbaum, Hugo Larochelle, and Richard~S. Zemel. 2018.
\newblock \href {https://openreview.net/forum?id=HJcSzz-CZ} {Meta-learning for
  semi-supervised few-shot classification}.
\newblock In \emph{International Conference on Learning Representations}.

\bibitem[{Snell et~al.(2017)Snell, Swersky, and Zemel}]{NIPS2017_6996}
Jake Snell, Kevin Swersky, and Richard Zemel. 2017.
\newblock \href
  {http://papers.nips.cc/paper/6996-prototypical-networks-for-few-shot-learning.pdf}
  {Prototypical networks for few-shot learning}.
\newblock In I.~Guyon, U.~V. Luxburg, S.~Bengio, H.~Wallach, R.~Fergus,
  S.~Vishwanathan, and R.~Garnett, editors, \emph{Advances in Neural
  Information Processing Systems 30}, pages 4077--4087. Curran Associates, Inc.

\bibitem[{Sun et~al.(2019)Sun, Sun, Zhou, and Lv}]{sun-etal-2019-hierarchical}
Shengli Sun, Qingfeng Sun, Kevin Zhou, and Tengchao Lv. 2019.
\newblock \href {https://doi.org/10.18653/v1/D19-1045} {Hierarchical attention
  prototypical networks for few-shot text classification}.
\newblock In \emph{Proceedings of the 2019 Conference on Empirical Methods in
  Natural Language Processing and the 9th International Joint Conference on
  Natural Language Processing (EMNLP-IJCNLP)}, pages 476--485, Hong Kong,
  China. Association for Computational Linguistics.

\bibitem[{Upadhyay et~al.(2018)Upadhyay, Faruqui, T{\"u}r, Dilek, and
  Heck}]{upadhyay2018almost}
Shyam Upadhyay, Manaal Faruqui, Gokhan T{\"u}r, Hakkani-T{\"u}r Dilek, and
  Larry Heck. 2018.
\newblock (almost) zero-shot cross-lingual spoken language understanding.
\newblock In \emph{2018 IEEE International Conference on Acoustics, Speech and
  Signal Processing (ICASSP)}, pages 6034--6038. IEEE.

\bibitem[{Vanschoren(2019)}]{Vanschoren2019}
Joaquin Vanschoren. 2019.
\newblock \href {https://doi.org/10.1007/978-3-030-05318-5_2}
  {\emph{Meta-Learning}}, pages 35--61. Springer International Publishing,
  Cham.

\bibitem[{Vaswani et~al.(2017)Vaswani, Shazeer, Parmar, Uszkoreit, Jones,
  Gomez, Kaiser, and Polosukhin}]{10.5555/3295222.3295349}
Ashish Vaswani, Noam Shazeer, Niki Parmar, Jakob Uszkoreit, Llion Jones,
  Aidan~N. Gomez, undefinedukasz Kaiser, and Illia Polosukhin. 2017.
\newblock Attention is all you need.
\newblock In \emph{Proceedings of the 31st International Conference on Neural
  Information Processing Systems}, NIPS’17, page 6000–6010, Red Hook, NY,
  USA. Curran Associates Inc.

\bibitem[{Vinyals et~al.(2016)Vinyals, Blundell, Lillicrap, Kavukcuoglu, and
  Wierstra}]{10.5555/3157382.3157504}
Oriol Vinyals, Charles Blundell, Timothy Lillicrap, Koray Kavukcuoglu, and Daan
  Wierstra. 2016.
\newblock Matching networks for one shot learning.
\newblock In \emph{Proceedings of the 30th International Conference on Neural
  Information Processing Systems}, NIPS’16, page 3637–3645, Red Hook, NY,
  USA. Curran Associates Inc.

\bibitem[{Wang et~al.(2019)Wang, Zheng, Yu, and Miao}]{10.1145/3293318}
Wei Wang, Vincent~W. Zheng, Han Yu, and Chunyan Miao. 2019.
\newblock \href {https://doi.org/10.1145/3293318} {A survey of zero-shot
  learning: Settings, methods, and applications}.
\newblock \emph{ACM Trans. Intell. Syst. Technol.}, 10(2).

\bibitem[{Wang et~al.(2018)Wang, Wu, Li, Gu, Xiang, Zhang, and
  Li}]{DBLP:journals/corr/abs-1807-02872}
Yong Wang, Xiao{-}Ming Wu, Qimai Li, Jiatao Gu, Wangmeng Xiang, Lei Zhang, and
  Victor O.~K. Li. 2018.
\newblock \href {http://arxiv.org/abs/1807.02872} {Large margin few-shot
  learning}.
\newblock \emph{CoRR}, abs/1807.02872.

\bibitem[{Xu et~al.(2020)Xu, Haider, and Mansour}]{xu-etal-2020-end}
Weijia Xu, Batool Haider, and Saab Mansour. 2020.
\newblock \href {https://doi.org/10.18653/v1/2020.emnlp-main.410} {End-to-end
  slot alignment and recognition for cross-lingual {NLU}}.
\newblock In \emph{Proceedings of the 2020 Conference on Empirical Methods in
  Natural Language Processing (EMNLP)}, pages 5052--5063, Online. Association
  for Computational Linguistics.

\bibitem[{Yazdani and Henderson(2015)}]{yazdani-henderson-2015-model}
Majid Yazdani and James Henderson. 2015.
\newblock \href {https://doi.org/10.18653/v1/D15-1027} {A model of zero-shot
  learning of spoken language understanding}.
\newblock In \emph{Proceedings of the 2015 Conference on Empirical Methods in
  Natural Language Processing}, pages 244--249, Lisbon, Portugal. Association
  for Computational Linguistics.

\bibitem[{Zhang et~al.(2020)Zhang, Williams, Titov, and
  Sennrich}]{zhang-etal-2020-improving}
Biao Zhang, Philip Williams, Ivan Titov, and Rico Sennrich. 2020.
\newblock \href {https://doi.org/10.18653/v1/2020.acl-main.148} {Improving
  massively multilingual neural machine translation and zero-shot translation}.
\newblock In \emph{Proceedings of the 58th Annual Meeting of the Association
  for Computational Linguistics}, pages 1628--1639, Online. Association for
  Computational Linguistics.

\bibitem[{Zhang et~al.(2019)Zhang, Li, Du, Fan, and Yu}]{zhang-etal-2019-joint}
Chenwei Zhang, Yaliang Li, Nan Du, Wei Fan, and Philip Yu. 2019.
\newblock \href {https://doi.org/10.18653/v1/P19-1519} {Joint slot filling and
  intent detection via capsule neural networks}.
\newblock In \emph{Proceedings of the 57th Annual Meeting of the Association
  for Computational Linguistics}, pages 5259--5267, Florence, Italy.
  Association for Computational Linguistics.

\bibitem[{Zhang and Wang(2016)}]{10.5555/3060832.3061040}
Xiaodong Zhang and Houfeng Wang. 2016.
\newblock A joint model of intent determination and slot filling for spoken
  language understanding.
\newblock In \emph{Proceedings of the Twenty-Fifth International Joint
  Conference on Artificial Intelligence}, IJCAI’16, page 2993–2999. AAAI
  Press.

\end{thebibliography}

%\appendix

\end{document}